%% file: main.tex
\documentclass[sigconf]{acmart}
\usepackage{algorithm}
\usepackage{algorithmicx, algpseudocode}
\usepackage{graphicx}
\usepackage{mathtools}
\usepackage{multirow}
\usepackage{subfigure}
\usepackage{mathtools}
\usepackage{thmtools}
\usepackage{rotating}
\usepackage{diagbox}
\usepackage{threeparttable}
\usepackage{color, colortbl}
\usepackage{amsmath}
\usepackage{arydshln}

\DeclareMathOperator{\argmax}{arg\,max}
\definecolor{LightCyan}{rgb}{0.88,1,1}
\newcommand{\tabincell}[2]{\begin{tabular}{@{}#1@{}}#2\end{tabular}}

\def \myalgPro {DGPN}

\AtBeginDocument{%
  \providecommand\BibTeX{{%
    \normalfont B\kern-0.5em{\scshape i\kern-0.25em b}\kern-0.8em\TeX}}}

\copyrightyear{2021}
\acmYear{2021}
\setcopyright{acmcopyright}\acmConference[KDD '21]{Proceedings of the 27th ACM
SIGKDD Conference on Knowledge Discovery and Data Mining}{August 14--18,
2021}{Virtual Event, Singapore}
\acmBooktitle{Proceedings of the 27th ACM SIGKDD Conference on Knowledge
Discovery and Data Mining (KDD '21), August 14--18, 2021, Virtual Event, Singapore}
\acmPrice{15.00}
\acmDOI{10.1145/3447548.3467230}
\acmISBN{978-1-4503-8332-5/21/08}

\acmSubmissionID{rst0238}

\begin{document}
\fancyhead{}
\title{Zero-shot Node Classification with Decomposed Graph Prototype Network}

\author{Zheng Wang$^{1, 2}$, Jialong Wang$^{1}$, Yuchen Guo$^{3}$, Zhiguo Gong$^{1*}$}
\affiliation{
    $^1$ State Key Laboratory of Internet of Things for Smart City, University of Macau
    \country{China}
}
\affiliation{
    $^2$ Department of Computer Science and Technology, University of Science and Technology Beijing
    \country{China}
}
\affiliation{
    $^3$ Institute for Brain and Cognitive Sciences, BNRist, Tsinghua University
    \country{China}
}
\email{zhengwang100@gmail.com, fstzgg@um.edu.mo}

\renewcommand{\shortauthors}{Z. Wang, et al.}

\begin{abstract}
\renewcommand{\thefootnote}{\fnsymbol{footnote}}
\footnotetext{* Zhiguo Gong is the corresponding author.}
\renewcommand{\thefootnote}{\arabic{footnote}}

Node classification is a central task in graph data analysis.
Scarce or even no labeled data of emerging classes is a big challenge for existing methods.
A natural question arises: can we classify the nodes from those classes that have never been seen?

In this paper, we study this zero-shot node classification (ZNC) problem which has a two-stage nature: (1) acquiring high-quality class semantic descriptions (CSDs) for knowledge transfer, and (2) designing a well generalized graph-based learning model.
For the first stage, we give a novel quantitative CSDs evaluation strategy based on estimating the real class relationships, to get the ``best'' CSDs in a completely automatic way.
For the second stage, we propose a novel Decomposed Graph Prototype Network (DGPN) method, following the principles of locality and compositionality for zero-shot model generalization.
Finally, we conduct extensive experiments to demonstrate the effectiveness of our solutions.
\end{abstract}



\maketitle
\input{1.intro}
\input{2.related}

\input{3.problem}
\input{4.csds}
\input{5.method}

\input{6.exp}
\input{7.conclusion}

\begin{acks}
This work is supported by National Natural Science Foundation of China (61902020), National Key D\&R Program of China (2019YFB1600704), Macao Youth Scholars Program (AM201912), The Science and Technology Development Fund, Macau SAR (0068/2020/AGJ, 0045/2019/A1, 0007/2018/A1, SKL-IOTSC-2021-2023), GSTIC (201907010013, EF005/FST-GZG/2019/GSTIC), University of Macau (MYRG2018-00129-FST), and Fundamental Research Funds for the Central Universities (FRF-TP-20-040A2).
\end{acks}

\bibliographystyle{ACM-Reference-Format}
\bibliography{sample-base}

\appendix
\input{8.appendix}

\end{document}

%% file: 1.intro.tex
\section{Introduction}
Node classification is an integral component of graph data analysis~\cite{cook2006mining,wang2016causality}, like document classification in citation networks~\cite{sen2008collective}, user type prediction in social networks~\cite{jin2013understanding}, and protein function identification in bioinformatics~\cite{kanehisa2003bioinformatics}.
In a broadly applicable scenario, given a graph in which only a small fraction of nodes are labeled, the goal is to predict the labels of the remaining unlabeled ones, assuming that graph structure information reflects some affinities among nodes.
This problem is well-studied, and solutions are often powerful~\cite{chapelle2009semi}.

\begin{figure}[!t]
\centering
    \includegraphics[width=0.5\textwidth]{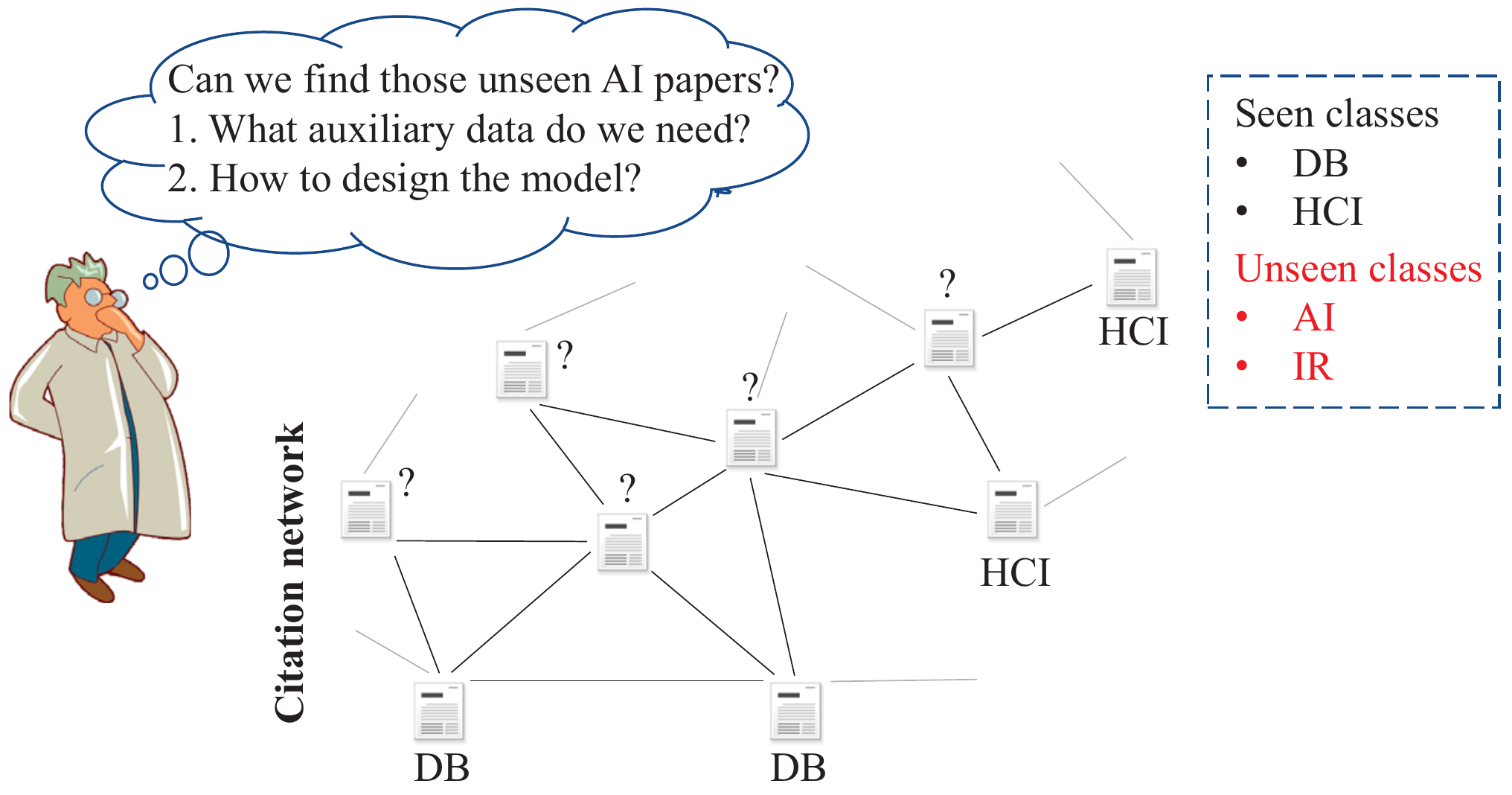}
\caption{
The problem of zero-shot node classification.}
\label{fig_problem}
\end{figure}

\subsubsection*{\textbf{Problem}}
Traditional node classification methods are facing an enormous challenge --- cannot catch up with the rapid growth of newly emerging classes in dynamic and open environments.
For example, it is hard and costly to annotate sufficient labeled nodes for a new research topic in citation networks; moreover and in fact, it is impossible to know the exact class numbers in real-world big data applications (like Wikipedia).
Naturally, as illustrated in Figure~\ref{fig_problem}, it would be very useful to empower models the ability to classify the nodes from those ``unseen'' classes that have no labeled instances.

This paper presents the first study of this zero-shot node classification (ZNC) problem.
As a practical application of zero-shot learning (ZSL)~\cite{farhadi2009describing}~\cite{larochelle2008zero}, this problem has a two-stage nature.
The first refers to acquiring high-quality class semantic descriptions (CSDs)\footnote{In computer vision, this kind of semantic knowledge is usually referred as ``attributes''. In this paper, to avoid ambiguity with node attributes (i.e., original features like bag-of-words in documents) terminology, we use CSDs instead.} as auxiliary data, for transferring supervised knowledge from seen classes to unseen classes.
However, few studies have explored this kind of semantic knowledge in graph scenarios where classes (e.g., ``AI'' papers in citation networks) are generally complex and abstract.
Moreover, the study of ``what are good CSDs?'' is also very limited.

The other stage refers to designing a well-generalized graph-based learning model.
Unlike the traditional ZSL tasks, node classification is a type of relational learning task that operates on graph-structured data.
Moreover, how to effectively utilize the rich graph information (like node attributes and graph structure) together with the CSDs knowledge for zero-shot models is still an open problem.

\subsubsection*{\textbf{Solutions and Contributions}}
For the first stage, considering the difficulty of manually describing complex and abstract concepts in graph scenarios, we propose to acquire high-quality CSDs in a completely automatic way.
For illustration, we use the famous natural language processing model Bert-Tiny~\cite{turc2019well} to automatically generate two types of CSDs: LABEL-CSDs (i.e., the word embeddings generated from class labels/names) and TEXT-CSDs (i.e., the document embeddings generated from some class-related text descriptions).
Next, as it is inconvenient for humans to incorporate domain knowledge in the learned embedding space, we give a novel quantitative CSDs evaluation strategy, based on the assumption that good CSDs should reflect the real relationships among classes~\cite{zhang2015zero}.
As such, we can finally choose the ``best'' CSDs from the above two kinds of candidates.

For the second stage, we propose a novel Decomposed Graph Prototype Network (DGPN) method, following the principles of locality (i.e., the concern about the small subparts of input) and compositionality (i.e., the concern about the combination of representations learned from these small subparts) for zero-shot model generalization~\cite{sylvain2019locality}.
In particular, we first show how to decompose the outputs of multi-layer graph convolutional networks (GCNs)~\cite{kipf2016semi}.
Then, for locality, we inject some semantic losses on those intermediate representations learned from these decompositions.
Finally, for compositionality, we apply a weighted sum pooling operation to those intermediate representations to get global ones for the target problem.
Intuitively, our method not only enhances the locality of node representations that is critical for zero-shot generalization, but also guarantees the discriminability of the global compositional representation for the final node classification problem.

We experimentally demonstrate the effectiveness of the proposed CSDs evaluation strategy as well as the proposed method DGPN, from which we can get some interesting findings.
Firstly, the quality of CSDs is the key to the ZNC problem.
More specifically, comparing to graph information, we can rank their importance as: CSDs $\gg$ node attributes $>$ graph structure.
Secondly, comparing to the performance of the most na\"ive baseline ``RandomGuess'', we can rank the general performance of those two CSDs as: TEXT-CSDs $\gg$ LABEL-CSDs $\ge$ RandomGuess.
Thirdly, with high-quality CSDs, graph structure information can be very useful or even be comparable to node attributes.
Lastly, through subtly recasting the concepts, the principles of locality and compositionality can be well adapted to graph-structured data.

The main contributions of this paper are summarized as follows:
\begin{itemize}
  \item \emph{Novel problem and findings}. We study the problem of zero-shot node classification.
          Through various experiments and analysis, we uncover some new and interesting findings of ZSL in the graph scenario.
  \item \emph{Novel CSDs acquisition and evaluation strategy}. We give a novel quantitative CSDs evaluation strategy, so as to acquire high-quality CSDs in a completely automatic way.
  \item \emph{Novel ZNC method}. We propose \myalgPro\ for the studied ZNC problem, following the principles of locality and compositionality for zero-shot model generalization.
\end{itemize}

\subsubsection*{\textbf{Roadmap}}
The remainder of this paper is organized as follows.
We review some related work in Section~\ref{sect_related}, and discuss the studied problem in Section~\ref{sect_problem}.
In Section~\ref{sect_CSD}, by introducing a novel quantitative CSDs evaluation strategy, we show how to acquire the ``best'' CSDs in a completely automatic way.
In Section~\ref{sect_method}, we elaborate the proposed method \myalgPro\ in details.
In Section~\ref{sect_exp}, we report the comparison of our method with existing ZSL methods.
Finally, we conclude this paper in Section~\ref{sect_conclusion}.

%% file: 2.related.tex
\section{Related Work}\label{sect_related}
\subsection{Node Classification}
Early studies~\cite{zhou2004learning}~\cite{zhu2003semi} generally use shallow models to jointly consider the graph structure and supervised information.
With the recent dramatic progress in deep neural networks, graph neural network (GNN)~\cite{wu2020comprehensive} methods are becoming the primary techniques.
Generally, GNN methods stack multiple neural network layers to capture graph information, and end with a classification layer to utilize the supervised knowledge.
Specifically, at each layer, GNN methods propagate information from each node to its neighborhoods with some well-defined propagation rules.
The most well-known work is GCN~\cite{kipf2016semi} which propagates node information based on a normalized and self-looped adjacency matrix.
Recent attempts to advance this line are GAT~\cite{velickovic_2018_iclr}, LNGN~\cite{de2020natural} and so on.

Nevertheless, existing methods generally all assume that every class in the graph has some labeled nodes.
The inability to generalize to unseen classes is one of the major challenges for current methods.

\subsection{Zero-shot Learning}\label{subsect_zsl}
Zero-shot learning (ZSL)~\cite{farhadi2009describing} (also known as zero-data learning~\cite{larochelle2008zero}) has recently become
a hot topic in machine learning and computer vision areas.
The goal is to classify the samples belonging to the classes which have no labeled data.
To solve this problem, class semantic descriptions (CSDs)~\cite{lampert2013attribute}, which could enable cross-class knowledge transfer, are introduced.
For example, to distinguish animals, we can first define some CSDs like ``swim'', ``wing'' and ``fur''.
Then, at the training stage, we can train classifiers to recognize these CSDs.
At the testing stage, given an animal from some unseen classes, we can first infer its CSDs and then compare the results with those of each unseen class to predict its label~\cite{lampert2013attribute}~\cite{romera2015embarrassingly}.

However, existing ZSL methods are mainly limited to computer vision~\cite{wang2019survey} or natural language processing~\cite{yin2019benchmarking}.
Although, some studies~\cite{wang2018rsdne, wang2020RECT, wang2021expanding} also consider the zero-shot setting in graph scenarios, they focus on graph embedding~\cite{goyal2018graph, wang2017equivalence} not node classification.
Therefore, the graph scenario might become a new challenging application context for ZSL communities.

%% file: 3.problem.tex
\section{PROBLEM DEFINITION and DISCUSSION}\label{sect_problem}
\subsection{Problem Definition}
Let ${\mathcal{G}} = ({\mathcal{V}}, \mathcal{E})$ denote a graph, where $\mathcal{V}$ denotes the set of $n$ nodes $\{v_1, \dots, v_n\}$, and $\mathcal{E}$ denotes the set of edges between these $n$ nodes.
Let $ A\in\mathbb{R}^{n \times n}$ be the adjacency matrix with $A_{ij}$ denoting the edge weight between nodes $v_i$ and $v_j$.
Furthermore, we use $X \in \mathbb{R}^{n \times d}$ to denote the node attribute (i.e., feature) matrix, i.e., each node is associated with a $d$-dimensional attribute vector.
The class set in this graph is $\mathcal{C} = \{\mathcal{C}^{s} \cup \mathcal{C}^{u}\}$ where $\mathcal{C}^{s}$ is the seen class set and $\mathcal{C}^{u}$ is the unseen class set satisfying $\mathcal{C}^{s} \cap \mathcal{C}^{u} = \varnothing$.
Supposing all the labeled nodes are coming from seen classes, the goal of zero-shot node classification (ZNC) is to classify the rest testing nodes whose label set is $C^u$.

\subsection{Problem Discussion}
As a practical application of ZSL, the problem of ZNC has a two-stage nature.
The first and most important is: how to acquire high-quality CSDs which can be used as a bridge for cross-class knowledge transfer.
In other words, we want to (1) determine the way of obtaining CSDs, and (2) quantitatively measure the overall quality.
Existing ZSL methods, which are generally developed for computer vision, mainly rely on human annotations.
For example, given a class like ``duck'', they manually describe this class by some figurative words like ``wing'' and ``feather''.
However, this may not be practical for ZNC, as graphs generally have more complex and abstract classes.
For instance, in social networks, it is hard to figure out what a ``blue'' or ``optimistic'' user looks like; and in citation networks, it is hard to describe what are ``AI'' papers.
On the other hand, although there exist some automatic solutions, the related studies on graph scenarios are still unexplored.
Moreover, the limited study of quantitative CSDs evaluation further prevents their practical usage.

The other is: how to design a well-generalized graph-based learning model.
Specifically, we want to (1) effectively utilize the rich graph information (like node attributes and graph structure), and (2) make the model capable of zero-shot generalization.
Firstly, few ZSL methods have ever considered the graph-structured data.
On the other hand, with the given CSDs knowledge, how to design a graph-based learning model (especially a powerful GNN model) that generalizes well to unseen classes is still an open question.

In the following, we will elaborate our solutions by addressing these two subproblems sequentially. 

%% file: 4.csds.tex
\section{Acquiring High-quality CSD\lowercase{s}}\label{sect_CSD}
Considering the difficulty of manually describing complex and abstract concepts in graph scenarios, in this paper, we aim to acquire high-quality CSDs in a completely automatic way.
In this section, we first show how to get some CSDs candidates via a typical automatic tool, then give a novel quantitative CSDs evaluation strategy, and at last present a simple experiment to illustrate the whole acquisition process.

\subsection{Getting CSDs Candidates Automatically}\label{subsect_get_csd}
For the automatic acquisition of semantic knowledge, text is playing the most significant role at present.
The basic idea is that with the help of some machine-learning models, we can use the ``word2vec''~\cite{mikolov2013efficient} results of some related textural sources as CSDs.
Intuitively, in the learned word embedding space, each dimension denotes a ``latent'' unknown class property.
Recently, neural network models, like Bert~\cite{devlin2018bert}, have made remarkable achievements in lots of word embedding tasks.
For efficiency and ease of calculation, we use Bert-Tiny~\cite{turc2019well} (the light version of Bert) to generate the following two types of 128-dimensional CSDs:
\begin{enumerate}
  \item \emph{LABEL-CSDs} are the word/phrase embedding results generated from class labels, i.e., class names (like ``dog'' and ``cat'') usually in the form of a word or phrase.
  \item \emph{TEXT-CSDs} are the document embedding results generated from some class-related text descriptions (like a paragraph or an article describing this class). Specifically, for each class, we use its Wikipedia page as text descriptions.
\end{enumerate}

In the traditional ZSL application scenarios (like computer vision and natural language processing), LABEL-CSDs are usually used, due to the easy computation and good performance.
On the other hand, TEXT-CSDs usually suffer from heavy computation and huge noise~\cite{zhu2018generative}.

\subsection{Quantitative CSDs Evaluation}\label{subsect_csd_exps}
As it is inconvenient for humans to incorporate domain knowledge in the learned embedding space, traditional methods generally evaluate the quality of CSDs according to their performance on the target ZSL task.
However, this kind of ``ex-post forecast'' strategy seriously relies on the choice of some particular ZSL methods and the specific application scenarios.
In the following, we give a more practical and easily implemented evaluation strategy.

The evaluation is based on the assumption that good CSDs should reflect the real relationships among classes~\cite{zhang2015zero}.
Although those relationships are not explicitly given, we can approximately estimate them\footnote{This evaluation can be conducted on all classes or only on seen classes. In this paper, we choose the former one for a comprehensive investigation.}.
Specifically, we first use a set of mean-class feature vectors $\{o_{i}\}_{i=1}^{|\mathcal{C}|}$ (i.e., $o_i \in \mathbb{R}^{d}$ denotes the center of class $c_i\in \mathcal{C}$ in the $d$-dimensional feature space) as class prototypes, following the work in few-shot learning~\cite{snell2017prototypical}.
Then, for any two classes $c_i$ and $c_j$, we define the empirical probability of $c_j$ generated by $c_i$ as:
\begin{equation}\label{eq_softmax_empirical}
  Pr(c_j|c_i) = \frac{exp(o^\top_i \cdot o_j)}{\sum_{t, t\neq i}^{|C|}{exp(o^\top_i \cdot o_t)}}
\end{equation}
where $exp(\cdot)$ is the usual exponential function, and $\cdot$ stands for the inner product between two vectors.
Intuitively, for each class $c_i$, Eq.~\ref{eq_softmax_empirical} defines a conditional distribution $Pr(\cdot |c_i)$ over all the other classes.

Similarly, given the automatically generated CSD vectors of all classes $\{s_i\}_{i=1}^{|\mathcal{C}|}$ (i.e.,  $s_i \in \mathbb{R}^{d^s}$ denotes the $d^s$-dimensional CSD vector of class $c_i\in \mathcal{C}$) and a specific class pair $<c_i, c_j>$, we can also calculate the probability $\hat{P}r(c_j |c_i)$ as:
\begin{equation}\label{eq_softmax}
  \hat{P}r(c_j|c_i) = \frac{exp(s^\top_i \cdot s_j)}{\sum_{t, t\neq i}^{|C|}{exp(s^\top_i \cdot s_t)}}
\end{equation}

Finally, the quality of the generated CSDs can be evaluated by comparing the above two distributions:
\begin{equation}\label{eq_softmax_sum}
  \frac{1}{|\mathcal{C}|} \sum_{c_i \in \mathcal{C}} dis(Pr(\cdot| c_i), \hat{P}r(\cdot| c_i))
\end{equation}
where $dis(\cdot, \cdot)$ is the selected distance/similarity measure.
In this paper, we adopt three well-known measures: KL Divergence, Cosine Similarity and Euclidean Distance.

\begin{table}[!t]
\small
    \centering
    \caption{Quality of obtained CSDs.}
    \begin{tabular}{ll|ccc}
    Dataset& CSDs Type & \tabincell{c}{KL\\Divergence}$\downarrow$      &  \tabincell{c}{Cosine\\ Similarity}$\uparrow$    & \tabincell{c}{Euclidean\\ Distance}$\downarrow$      \\
    \midrule
    \multirow{2}{*}{Cora}
            & LABEL-CSDs & 0.0154& 0.9978& 0.1787 \\
            & TEXT-CSDs  &\textbf{0.0109}	&\textbf{0.9985}& \textbf{0.1552} \\
    \midrule
    \multirow{2}{*}{Citeseer}
            & LABEL-CSDs & 0.0120 & 0.9980 &	0.1620      \\
            & TEXT-CSDs & \textbf{0.0077} &	\textbf{0.9987} & \textbf{0.1328} \\
    \midrule
    \multirow{2}{*}{C-M10M}
    & LABEL-CSDs & 0.0062	& 0.9990& 0.1175       \\
    & TEXT-CSDs & \textbf{0.0026}   & \textbf{0.9996}   & \textbf{0.0735}   \\
    \bottomrule
    \end{tabular}
    \label{tab_csd_quality}
\begin{tablenotes}
  \item[*]\footnotesize Here, `$\downarrow$' indicates the lower the better, whereas `$\uparrow$' indicates the higher the better.
\end{tablenotes}
\end{table}
\subsection{CSDs Evaluation Experiment}\label{subsect_csd_eval_exp}
We conduct the evaluation on three real-world citation networks: Cora~\cite{sen2008collective}, Citeseer~\cite{sen2008collective}, and C-M10M~\cite{pan2016tri}.
The details about these datasets and this experiment can be found in Appendix~\ref{app_data_details} and Appendix~\ref{app_csds_exp_details}, respectively.
As shown in Table~\ref{tab_csd_quality}, TEXT-CSDs always get a much better performance.
This may due to the complex and abstract nature of graph classes, i.e., describing graph classes needs rich text (Wikipedia articles, in particular) which contains more elaborate and subtle information.
Therefore, we use TEXT-CSDs as auxiliary information in the later ZNC experiments. 

%% file: 5.method.tex
\section{Designing well-generalized graph-based learning models}\label{sect_method}
Recent studies~\cite{sylvain2019locality}~\cite{xu2020attribute} show that locality and compositionality are fundamental ingredients for well-generalized zero-shot models.
Intuitively, locality concerns the small subparts of input, and compositionality concerns the combination of representations learned from these small subparts.
In the following, we will first give a brief introduction to GCNs, then show how to decompose the outputs of GCNs into a set of ``subparts'', and finally elaborate the proposed method following the above two principles.

\begin{figure*}[!ht]\centering
    \centering
    \includegraphics[width=0.95\textwidth]{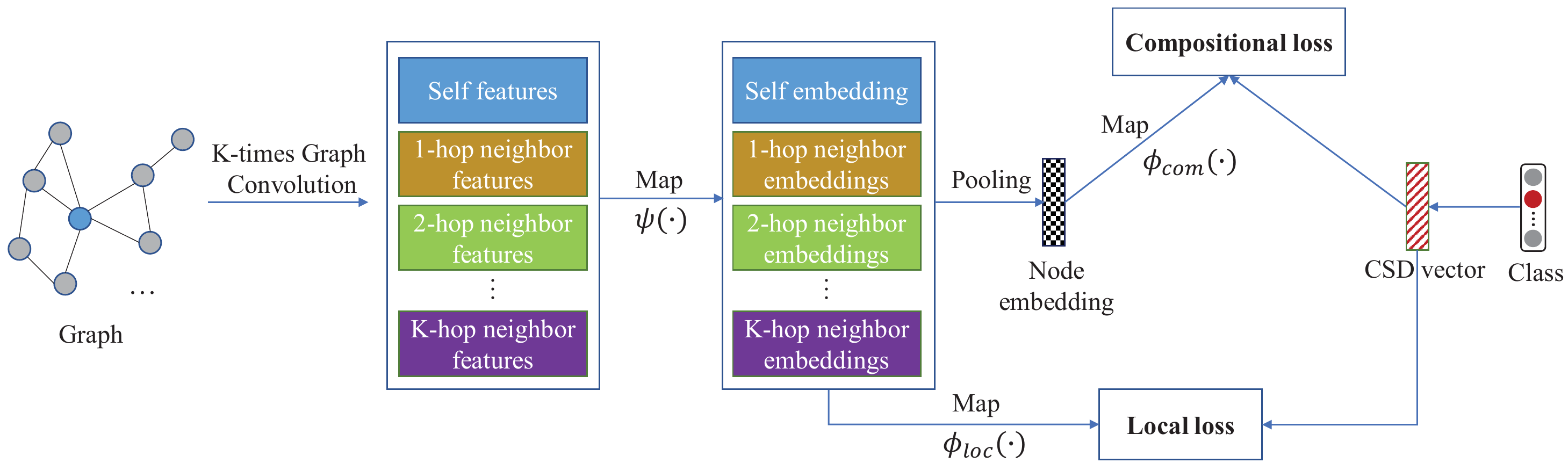}
    \caption{The architecture of Decomposed Graph Prototype Network (\myalgPro).}
    \label{fig_our_method}
\end{figure*}
\subsection{Preliminaries on GCNs}
We use the diagonal matrix $D$ to denote the degree matrix of the adjacency matrix A, i.e., $D_{ii} =  \sum_j A_{ij}$.
The normalized graph Laplacian matrix is $L=I_{n}-D^{-\frac{1}{2}}AD^{-\frac{1}{2}}$ which is a symmetric positive semidefinite matrix with eigen-decomposition $L = U\Lambda U^{\top}$.
Here $\Lambda$ is a diagonal matrix of the eigenvalues of $L$, and $U\in \mathbb{R}^{n \times n }$ is a unitary matrix which comprises orthonormal eigenvectors of $L$.

\subsubsection*{\textbf{Spectral Graph Convolution}}
We consider the graph convolution operation defined as the multiplication of a signal $x$ with a filter $g_{\theta}$ (parameterized by $\theta \in \mathbb{R}^{n}$) in the Fourier domain~\cite{shuman2013emerging}:
\begin{equation}\label{eq_gc_fd}
    g_{\theta} \star x=Ug_{\theta}(\Lambda)U^{\top}x
\end{equation}
where $\star$ denotes the convolution operator.
We can understand $g_{\theta}$ as a function operating on the eigenvalues of the laplacian matrix, i.e., $g_{\theta}(\Lambda)$.
For efficiency, $g_{\theta}(\Lambda)$ can be further approximated by the $K$-th order Chebyshev polynomials~\cite{hammond2011wavelets}~\cite{defferrard2016convolutional}, as follows:

\begin{equation}\label{eq_gc_fd_cheby}
   g_{\theta} \star x \approx U \sum_{k=0}^K \theta_k^\prime T_k(\tilde{\Lambda}) U^{\top}x = \sum_{k=0}^K \theta_k^\prime T_k(\tilde{L}) x
\end{equation}
where $T_k(\cdot)$ denotes the Chebyshev polynomials and $\theta^\prime$ denotes a vector of the Chebyshev coefficients;
$\tilde{\Lambda}=\frac{2}{\lambda_{max}}\Lambda-I_n$ is a rescaled diagonal eigenvalue matrix, $I_n$ is the identity matrix, $\lambda_{max}$ denotes the largest eigenvalue of $L$, and $\tilde{L}=U\tilde{\Lambda} U^{\top}=\frac{2}{\lambda_{max}}L-I_n$.

\subsubsection*{\textbf{Graph Convolutional Network (GCN)}}
The vanilla GCN~\cite{kipf2016semi} simplifies the Chebyshev approximation by setting $\lambda_{max} \approx 2$, $K=1$, and $\theta=\theta'_{0}=-\theta'_{1}$.
This leads to the following spectral convolution formulation:
\begin{equation}\label{eq_gc_fd_cheby_gcn}
    g_{\theta}\star x \approx \theta(I_{n} +D^{-\frac{1}{2}}AD^{-\frac{1}{2}})x
\end{equation}

Then it adopts a renormalization trick:
\begin{equation}\label{eq_gcn_trick}
  I_{n} +D^{-\frac{1}{2}}AD^{-\frac{1}{2}} \to \tilde{D}^{-\frac{1}{2}}\tilde{A}\tilde{D}^{-\frac{1}{2}}
\end{equation}
where $\tilde{A}=A+I_n$ and $\tilde{D}_{ii}=\sum_j\tilde{A}_{ij}$.
In this way, the spectral graph convolution is finally simplified as follows:
\begin{equation}\label{eq_gcn_trick_final}
    g_{\theta}\star x \approx \theta (\tilde{D}^{-\frac{1}{2}}\tilde{A}\tilde{D}^{-\frac{1}{2}})x
\end{equation}

\subsection{GCNs Decomposition}\label{subsect_decompose_gcns}
In computer vision where the concepts of locality and compositionality are initially introduced, a small subpart (usually known as ``patch'') of input usually refers to an image region, i.e., a small rectangular area of an image.
Generally, convolutional neural networks (CNNs)~\cite{krizhevsky2017imagenet}, which have a locality-aware architecture, can exploit this kind of local information from images intrinsically~\cite{sylvain2019locality}~\cite{xu2020attribute}.
However, CNNs cannot deal with the irregular graph structures which lack of shift-invariant notion of translation~\cite{shuman2012windowed}.
In the following, using two typical cases, we show how to decompose the outputs of GCNs into a set of subparts, so as to localize irregular graph structures.

\subsubsection{Case I: The Vanilla GCN without Renormalization Trick}\label{subsect_gcn_wo_renor}
We continue with the graph convolution formulated in Eq.~\ref{eq_gc_fd_cheby_gcn}.
In the following, we use $g_{\theta_k}$ to denote the filter (parameterized by $\theta_k$) at the $k$-th graph convolution, and let $P=D^{-\frac{1}{2}}AD^{-\frac{1}{2}}$.
As such, on signal $x$, performing the graph convolution $K$ times leads to:
\begin{equation}\label{eq_fourier_gcn_multiple_same_theta}
\begin{aligned}
  g_{\theta_K}\star \cdots g_{\theta_1} \star x & \approx \theta_{K}\cdots\theta_{1} (I_{n} +P)^{K}x\\
  & = \theta^*\sum_{k=0}^{K} \binom{K}{k} P^{k} x
\end{aligned}
\end{equation}
where $\theta^* = \theta_{K}\cdots\theta_{1} $ denotes the integration of all learnable parameters and $\binom{K}{k}$ is the combinatorial number.
However, Eq.~\ref{eq_fourier_gcn_multiple_same_theta} has the problem of numerical instability, as the eigenvalues of matrix $(I_{n} +P)$ are in the range $[0, 2]$ and those of $(I_{n} +P)^{K}$ are in the range $[0, 2^K]$.
To circumvent this problem, we can further normalize Eq.~\ref{eq_fourier_gcn_multiple_same_theta} by dividing by $2^K$, which leads:
\begin{equation}\label{eq_fourier_gcn_multiple_same_theta_normalized}
\begin{aligned}
  g_{\theta_K}\star \cdots \star (g_{\theta_1} \star x) & \approx \theta^*\sum_{k=0}^{K} \frac{\binom{K}{k}}{2^{K}} P^{k} x
\end{aligned}
\end{equation}

This indicates that the outputs of a $K$ times graph convolution over signal $x$ can be decomposed into a set of $K+1$ subparts (i.e., $\{P^{k} x\}_{k=0}^{K}$), with the $k$-th subpart reflecting the knowledge of its $k$-hop neighbors.

\subsubsection*{\textbf{The variant with lazy random walk}}
In order to alleviate the over-smoothing problem~\cite{li2018deeper} in GCNs, lazy random walk is usually considered.
Specifically, at each graph convolution layer, it enables each node to maintain some parts of its representations learned in the preceding layer.

In the following, we continue with the graph convolution formulated in Eq.~\ref{eq_gc_fd_cheby}, and set $K=1$ and $\theta=\theta'_{0}=-\theta'_{1}$.
As such, we can get the following formulation:
\begin{equation}\label{eq_fourier_gcn_basic_gamma}
g_{\theta} \star x \approx \theta( (2-\frac{2}{\lambda_{max}})I_{n} + \frac{2}{\lambda_{max}} P)x
\end{equation}

As the matrix $((2-\frac{2}{\lambda_{max}})I_{n} + \frac{2}{\lambda_{max}} P)$ has eigenvalues in the range $[0, 2]$, for numerical stability, we further normalize it by dividing by $2$. This leads to the following equation:
\begin{equation}\label{eq_fourier_gcn_basic_gamma_normalized}
\begin{aligned}
 g_{\theta} \star x & \approx \theta(\beta I_{n} + (1-\beta) P)x
\end{aligned}
\end{equation}
where $\beta = 1-\frac{1}{\lambda_{max}}$.
If we enforce $\beta$ to lie in $[0,1]$, $\beta$ can be seen as the probability of staying at the current node in a lazy random walk.
This shows that compared to the vanilla GCN, lazy random walk is also a kind of spectral graph convolution model but with a different approximation for $\lambda_{max}$ (i.e., $\lambda_{max} > 2$) in the Chebyshev polynomials.

Based on Eq.~\ref{eq_fourier_gcn_basic_gamma_normalized}, on signal $x$, performing the lazy random walk $K$ times leads to:
\begin{equation}\label{eq_fourier_gcn_multiple_same_theta_lrw}
\begin{aligned}
  g_{\theta_K}\star \cdots \star (g_{\theta_1} \star x) & \approx \theta^*\sum_{k=0}^{K} \binom{K}{k} \beta ^{K-k} [(1-\beta) P]^{k}x
\end{aligned}
\end{equation}

\subsubsection{Case II: The Vanilla GCN with Renormalization Trick}\label{subsubsect_gcn_de_trick}
We continue with the trick formulated in Eq.~\ref{eq_gcn_trick}:
\begin{equation}\label{eq_fourier_gcn_multiple_different_theta}
\begin{aligned}
  I_{n} +D^{-\frac{1}{2}}AD^{-\frac{1}{2}} \to \tilde{D}^{-\frac{1}{2}}\tilde{A}\tilde{D}^{-\frac{1}{2}} = \tilde{I}_{n} + \tilde{P}
\end{aligned}
\end{equation}
where $\tilde{I}_{n}=\tilde{D}^{-\frac{1}{2}}I_{n}\tilde{D}^{-\frac{1}{2}}$ and $\tilde{P} = \tilde{D}^{-\frac{1}{2}}A\tilde{D}^{-\frac{1}{2}}$.

In this case, on signal $x$, performing the graph convolution (formulated in Eq.~\ref{eq_gcn_trick_final}) $K$ times leads to:
\begin{equation}\label{eq_vanila_gcn_trick_decompose_final}
\begin{aligned}
  g_{\theta_K}\star \cdots \star (g_{\theta_1} \star x) & \approx \theta^*(\tilde{I}_{n} +\tilde{P})^{K}x\\
  & = \theta^*\sum_{k=0}^{K} \binom{K}{k} \tilde{I}_{n}^{K-k}  \tilde{P}^{k}x
\end{aligned}
\end{equation}

\subsubsection*{\textbf{The variant with lazy random walk}} Similar to the analysis in Section~\ref{subsect_gcn_wo_renor}, we can further introduce lazy random walk into the above graph convolution.
We neglect these derivation processes, due to space limitation.
Finally, we can get:
\begin{equation}\label{eq_vanila_gcn_multiple_same_theta_lzw}
\begin{aligned}
  g_{\theta_K}\star \cdots \star (g_{\theta_1} \star x) & \approx \theta^*\sum_{k=0}^{K} \binom{K}{k} \beta ^{K-k} [(1-\beta) \hat{P}]^{k}x
\end{aligned}
\end{equation}
where $\hat{P} = \tilde{D}^{-\frac{1}{2}}\tilde{A}\tilde{D}^{-\frac{1}{2}}$.

\subsubsection{Time Complexity of GCNs Decomposition}\label{subsect_tc_de_gcns}
Without loss of generality, we continue with the Case I in Section~\ref{subsect_gcn_wo_renor}.
As shown in its formulation (Eq.~\ref{eq_fourier_gcn_multiple_same_theta_normalized}), we further get another $K$ new feature matrices; and each of them will cost $O(|\mathcal{E}|d)$, where $|\mathcal{E}|$ is the edge number.
Therefore, as a whole, the overall time complexity is $O(K|\mathcal{E}|d)$.

\subsection{Locality and Compositionality in GCNs}
Figure~\ref{fig_our_method} illustrates the proposed \myalgPro\ method for the studied ZNC problem.
In the following, we will show the design details by elaborating how to improve locality and compositionality in GCNs.
A schematic explanation of these two concepts can be found in Figure~\ref{fig_local_global} in Appendix~\ref{app_explain_lg}.

\subsubsection{Locality in GCNs}
From the viewpoint of representation learning, locality refers to the ability of a representation to localize and associate an input ``patch'' with semantic knowledge~\cite{sylvain2019locality}.
As analysed in~\ref{subsect_decompose_gcns}, the outputs of a $K$ times graph convolution over a node can be decomposed into $K+1$ subparts; each subpart can be seen as a ``patch'' containing the knowledge of a fixed hop of neighbors.
To improve locality, we can inject some semantic losses on those intermediate representations learned from these subparts.

\begin{table*}[!t]
\caption{Summary of the datasets.}
\small
\label{tab_datasets}
\begin{center}
    \begin{tabular}{l c c c c c c}
    \toprule
    {\bf Dataset} & {\bf Nodes} & {\bf Edges} & {\bf Features} &  {\bf Classes} & {\bf Class Split I [Train/Val/Test]} & {\bf Class Split II [Train/Val/Test]} \\ \midrule
    {Cora}  & 2,708 & 5,429 & 1,433 & 7 &[3/0/4]  & [2/2/3] \\
    {Citeseer}  & 3,327 & 4,732 & 3,703 & 6 & [2/0/4] & [2/2/2] \\
    {C-M10M} &4,464 &5,804 &128  & 6 &[3/0/3]  & [2/2/2] \\
    \bottomrule
    \end{tabular}
    \label{tab_sod}
\end{center}
\end{table*}

Without loss of generality, we use the decomposition of the vanilla GCN (formulated in Eq.~\ref{eq_fourier_gcn_multiple_same_theta_normalized}) as an example.
First of all, considering a $K$ times graph convolution, we can define the resulted node representations as the collection of $K+1$ subparts: $\{ P^{k} X \}_{k=0}^{K}$.
In other words, the resulted node $v_i$'s representation vector can be defined as the collection of itself and its $\{1,..., K\}$-hop neighbor information: $\{ \bar{x}^{(k)}_{i} \}_{k=0}^{K}$, where $\bar{x}^{(k)}_{i}$ denotes the $i$-th row of matrix $(P^{k} X)$.

With an input subpart $\bar{x}^{(k)}_{i}$, we can adopt a map function $\psi(\cdot)$ to convert it into a latent feature representation $h^{(k)}_{i} = \psi(\bar{x}^{(k)}_{i})$.
Then, we can force the resulted representation to encode some semantic knowledge.
In detail, we can use another map function $\phi_{loc}(\cdot)$ to map $h^{(k)}_{i}$ into a semantic space.
In this space, we can compute the prediction score $f^{(k)}_{ic}$ of node $v_i$ w.r.t. a seen class $c \in \mathcal{C}^{s}$ as:
\begin{equation}\label{eq_sim_semantic}
\begin{aligned}
f^{(k)}_{ic} = \mathrm{sim}(\phi_{loc}(h^{(k)}_{i}), s_c)
\end{aligned}
\end{equation}
where $\mathrm{sim}(\cdot, \cdot)$ is a similarity measure function (e.g., inner product and cosine similarity), and $s_c$ is the given CSD vector of class $c$.
It is worth noting that, for each node $v_i$, we obtain all its sub representations $\{h^{(k)}_{i}\}_{k=0}^{K}$ (via the map function $\psi(\cdot)$) at the same layer.
This would facilitate the configuration of local loss, i.e., avoiding the confusion of choosing appropriate intermediate layers in deep neural network models.

Next, we can apply a softmax function to transform the predicted scores into a probability distribution over all source classes.
Finally, the model is trained to minimize the cross-entropy loss between the predicted probabilities and the ground-truth labels.
Specifically, given a training node $v_i$ from a seen class $c \in \mathcal{C}^{s}$, the local loss (w.r.t. all its $K+1$ sub-representations) can be calculated as:
\begin{equation}\label{eq_local_loss}
\begin{aligned}
\mathcal{Q}_{loc} =- \sum_{k=0:K} \ln \frac{exp(f^{(k)}_{ic})}{\sum_{c'\in \mathcal{C}^{s}} exp(f^{(k)}_{ic'}) } \\
\end{aligned}
\end{equation}

Intuitively, by minimizing the above objective function, for each node, we can force the first neural network module (i.e., the map function $\psi(\cdot)$) in our method to extract a set of semantically relevant representations.

\subsubsection{Compositionality for GCNs}
From the viewpoint of representation learning, compositionality refers to the ability to express the learned global representations as a combination of those pre-learned sub-representations~\cite{andreas2018measuring}.
Based on the analysis in Section~\ref{subsect_decompose_gcns}, for each node $v_i$, we can apply a global weighted sum pooling operation on its previously learned $K+1$ sub-representations $\{h^{(k)}_{i}\}_{k=0}^{K}$, so as to obtain a global representation $z_i$:
\begin{equation}\label{eq_xxx}
z_i = \sum_{k=0}^K  \omega_{k} * h^{(k)}_{i}
\end{equation}
where $\{\omega_{k}\}_{k=0}^{K}$ are the scalar weight parameters, and symbol $*$ stands for the scalar multiplication.
This weight parameter vector is determined based on the decomposition analysis, e.g., $\omega_{k} = \frac{\binom{K}{k}}{2^{K}} $ in Case I (formulated in Eq.~\ref{eq_fourier_gcn_multiple_same_theta_normalized}).

For ZSL, we can also minimize a cross-entropy loss function in the semantic space.
Specifically, given a training node $v_i$ from a seen class $c \in \mathcal{C}^{s}$, the compositional loss can be calculated as:
\begin{equation}\label{eq_compositional_loss}
\mathcal{Q}_{com} = -\ln \frac{exp(g_{ic})}{\sum_{c'\in \mathcal{C}^{s}} exp(g_{ic'}) }
\end{equation}
where $g_{ic} = \mathrm{sim}(\phi_{com}(z_i), s_c)$ is the predicted score of node $v_i$ w.r.t. the seen class $c$ in the semantic space; and $\phi_{com}(\cdot)$ is a map function that maps the global feature $z_i$ into this semantic space.

\subsubsection{Joint Locality and Compositionality Graph Learning}
As illustrated in Figure~\ref{fig_our_method}, our full model \myalgPro\ optimizes the neural networks by integrating both the compositional loss (Eq.~\ref{eq_compositional_loss}) and local loss (Eq.~\ref{eq_local_loss}):
\begin{equation}\label{eq_loss_final}
\mathcal{Q} = \mathcal{Q}_{com} + \alpha \mathcal{Q}_{loc}
\end{equation}
where $\alpha$ is a hyper-parameter.
This joint learning not only enhances the locality of the node representation that is critical for zero-shot generalization, but also guarantees the discriminability of the global compositional representation for the final node classification.

After model convergence, given a node $v_i$ from unseen classes, we can infer its label from the unseen class set $\mathcal{C}^u$ as:
\begin{equation}\label{eq_infer}
\begin{aligned}
\argmax_{c\in \mathcal{C}^u} \mathrm{sim}(\phi_{com}(z_i), s_c)
\end{aligned}
\end{equation}
It is worth noting that, by changing the convolution time $K$ and choosing simple map functions for $\psi(\cdot)$, $\phi_{com}(\cdot)$ and $\phi_{loc}(\cdot)$, our method can preserve the high-order proximity of a graph, using only a few parameters.

\subsubsection{Time Complexity}
Suppose we adopt single-layer perceptrons for all these three map functions $\psi(\cdot)$, $\phi_{com}(\cdot)$ and $\phi_{loc}(\cdot)$.
First of all, as analysed in Section~\ref{subsect_tc_de_gcns}, the decomposition will cost $O(K|\mathcal{E}|d)$.
Then, all these K+1 subparts will be mapped to a $d^{h}$-dimensional hidden space, which will cost $O( (K+1) ndd^{h})$.
The afterwards pooling operator will cost $O(K n d^{h})$.
At last, all the intermediate results will be finally mapped to a $d^{s}$-dimensional semantic space, the time complexity of which would be $O((K+2)n d^{h}d^{s})$.
As a whole, the computational complexity of evaluating Eq.~\ref{eq_loss_final} is $O(K|\mathcal{E}|d d^{h} + Kn d^{h}d^{s})$, i.e., linear in the number of graph edges and nodes.

%% file: 6.exp.tex
\section{Experiment}\label{sect_exp}
In this section, we conduct a set of experiments to answer the following research questions:

\begin{itemize}
  \item \emph{RQ1:} Is it possible to conduct ZSL on graph-structured data? Especially, does the proposed method \myalgPro\ significantly outperform state-of-the-art ZSL methods?
  \item \emph{RQ2:} Which parts really affect the performance of \myalgPro? Or more subtly, the quality of CSDs, the used graph structure information, or the employed algorithm components?
  \item \emph{RQ3:} Can the decomposed GCNs part in \myalgPro\ be used for other applications?
\end{itemize}

\subsection{Experimental Setup}\label{subsect_exp_setup}
\subsubsection*{\textbf{Datasets}}
As summarized in Table~\ref{tab_datasets}, we use three widely used real-world citation networks: Cora~\cite{sen2008collective}, Citeseer~\cite{sen2008collective}, and C-M10M (a light version of Citeseer-M10~\cite{pan2016tri}).
In these datasets, nodes are publications, and edges are citation links; each node is associated with an attribute vector and belongs to one of the research topics.
To construct zero-shot setting, we design two fixed seen/unseen class split settings, for ease of comparison.
Specifically, based on their class IDs in each dataset, we adopt the first few classes as seen classes and the rest classes as unseen ones:
\begin{itemize}
  \item \emph{Class Split I}: all the seen classes are used for training, and all the unseen classes are used for testing.
  \item \emph{Class Split II}: the seen classes are further partitioned to train and validation parts, and all the unseen classes are still used for testing.
\end{itemize}
As analysed in Section~\ref{sect_CSD}, by default, we adopt the 128-dimensional TEXT-CSDs generated by Bert-Tiny as auxiliary data.
Details about these datasets and seen/unseen class split settings can be found in Appendix~\ref{app_data_details}.

\subsubsection*{\textbf{Baselines}}
The compared baselines include both classical and recent state-of-the-art ZSL methods: \emph{DAP}~\cite{lampert2013attribute}, \emph{ESZSL}~\cite{romera2015embarrassingly}, \emph{ZS-GCN}~\cite{wang2018zero}, \emph{WDVSc}~\cite{wan2019transductive}, and \emph{Hyperbolic-ZSL}~\cite{liu2020hyperbolic}.
In addition, as traditional methods are mainly designed for computer vision, their original implementations heavily rely on some pre-trained CNNs.
Therefore, we further test two representative variants: \emph{DAP(CNN)} and \emph{ZS-GCN(CNN)}, in both of which a pre-trained AlexNet~\cite{krizhevsky2017imagenet} CNN model is used as the backbone network.
Besides, \emph{RandomGuess} (i.e., randomly guessing an unseen label) is introduced as the na\"ive baseline.

\subsubsection*{\textbf{Parameter Settings}}
In our method, we use the decomposition of the vanilla GCN with lazy random walk (formulated in Eq.\ref{eq_fourier_gcn_multiple_same_theta_lrw}), employ single-layer perceptrons for all three map functions, and adopt the inner product as the similarity function.
At the first layer, the input size is equal to feature dimension, and the output size is simply fixed to 128.
At the second layer, the output dimension size is also set to 128, so as to be compatible with the given TEXT-CSDs for the final loss calculation.
Unless otherwise noted, all these settings are fixed throughout the whole experiment.

In addition, in Class Split I, we adopt the default hyper-parameter settings for all baselines.
For our method, we simply fix $K=3$ and $\beta=0.7$ in all datasets.
In Class Split II, the hyper-parameters in baselines and ours are all determined based on their performance on
validation data.
More details about these baselines and hyper-parameter settings can be found in Appendix~\ref{app_baselines}.

\begin{table}[!t]
\small
\caption{Zero-shot node classification accuracy (\%).}
\centering
\begin{tabular}{ll|ccc}
\toprule
& & Cora & Citeseer & C-M10M  \\
\hline
\multirow{10}{*}{\begin{turn}{90}\textbf{Class Split I}\end{turn}}
&RandomGuess	&25.35$\pm$1.28	&24.86$\pm$1.63	&33.21$\pm$1.08\\
&DAP	&26.56$\pm$0.37	&\underline{34.01$\pm$0.97}	&\underline{38.71$\pm$0.54}\\
&DAP(CNN)	&27.80$\pm$0.67	&30.45$\pm$0.93	&32.97$\pm$0.71\\
&ESZSL	&27.35$\pm$0.00	&30.32$\pm$0.00	&37.00$\pm$0.00\\
&ZS-GCN	&25.73$\pm$0.46	&28.62$\pm$0.20	&37.89$\pm$1.15\\
&ZS-GCN(CNN)	&16.01$\pm$3.27	&21.18$\pm$1.58	&36.44$\pm$0.97\\
&WDVSc	&\underline{30.62$\pm$0.38}	&23.46$\pm$0.11	&38.12$\pm$0.35\\
&Hyperbolic-ZSL	&26.36$\pm$0.41	&34.18$\pm$0.88	&35.80$\pm$2.25\\
&\cellcolor{LightCyan}\myalgPro\ (ours)	&\cellcolor{LightCyan}\textbf{33.78$\pm$0.28}	&\cellcolor{LightCyan}\textbf{38.02$\pm$0.11}	&\cellcolor{LightCyan}\textbf{41.98$\pm$0.21}\\
\cdashline{2-5}
& Improve$\uparrow$ & +10.32\% &+11.79\% &+8.45\% \\
\midrule
\multirow{10}{*}{\begin{turn}{90}\textbf{Class Split II}\end{turn}}
&RandomGuess	&32.69$\pm$1.48	&50.48$\pm$1.70	&49.73$\pm$1.56\\
&DAP	&30.22$\pm$1.21	&53.30$\pm$0.22	&46.79$\pm$4.16\\
&DAP(CNN)	&29.83$\pm$1.23	&50.07$\pm$1.70	&46.29$\pm$0.36\\
&ESZSL	&\underline{38.82$\pm$0.00}	&\underline{55.32$\pm$0.00}	&\underline{56.07$\pm$0.00}\\
&ZS-GCN	&29.53$\pm$0.91	&52.22$\pm$1.21	&55.28$\pm$0.41\\
&ZS-GCN(CNN)	&33.20$\pm$0.32	&49.27$\pm$0.73	&51.37$\pm$1.27\\
&WDVSc	&34.13$\pm$0.67	&52.70$\pm$0.68	&46.26$\pm$2.58\\
&Hyperbolic-ZSL	&37.02$\pm$0.28	&46.27$\pm$0.39	&55.07$\pm$0.77\\
&\cellcolor{LightCyan}\myalgPro\ (ours)	&\cellcolor{LightCyan}\textbf{46.40$\pm$0.31}	&\cellcolor{LightCyan}\textbf{61.90$\pm$0.32}	&\cellcolor{LightCyan}\textbf{62.46$\pm$0.42}\\
\cdashline{2-5}
& Improve$\uparrow$ & +19.53\%  & +11.89\% &+11.40\% \\
\bottomrule
\end{tabular}
\begin{tablenotes}
  \item[*]\footnotesize The best method is bolded, and the second-best is underlined.
\end{tablenotes}
\label{tab_zsl_acc}
\end{table}

\subsection{Over-all Performance (RQ1)}\label{subsect_exp_znc}
Table~\ref{tab_zsl_acc} shows the comparison results.
Firstly, we can see that our method \myalgPro\ always outperforms all baselines by a significant margin.
Compared to the best baseline, our method on average gives 10.19\% and 14.27\% improvements under the settings of Class Split I and Class Split II, respectively.
Secondly, although all baselines perform poorly on the whole, most of them still outperform RandomGuess.
Finally, the performance of both DAP and ZS-GCN almost always becomes worse when the pre-trained AlexNet model is involved.
Even more surprisingly, those simple classical methods (like DAP and ESZSL) generally get better results than those recently proposed complex ones (like ZS-GCN and Hyperbolic-ZSL).
This indicates that as a new problem, ZNC would become a new challenge for ZSL and graph learning communities.

\emph{Overall, the above experiments show the feasibility of conducting ZSL on graph-structured data.
In addition, for this new problem, our method is more effective than those traditional ZSL methods.}
\begin{table}[!t]
\setlength{\tabcolsep}{4pt} 
\small
\caption{Zero-shot node classification accuracy (\%) using LABEL-CSDs. }
\centering
\begin{tabular}{ll|rr|rr|rr}
\toprule
& &\multicolumn{2}{c|}{Cora} &\multicolumn{2}{c|}{Citeseer} &\multicolumn{2}{c}{C-M10M}\\
& & Acc. & Decl. & Acc. & Decl. & Acc. & Decl.  \\
\hline
\multirow{6}{*}{\begin{turn}{90}\textbf{Class Split I}\end{turn}}
&DAP	&25.34 &-4.59\%	&\textcolor[rgb]{0.00,0.00,1.00}{30.01} &-11.76\%	&32.67 &-15.60\%	\\
&ESZSL	&\textcolor[rgb]{0.00,0.00,1.00}{25.79} &-5.70\%	&\textcolor[rgb]{0.00,0.00,1.00}{28.52} &-5.94\%	&\textcolor[rgb]{0.00,0.00,1.00}{35.02} &-5.35\%	\\
&ZS-GCN	&23.73 &-7.77\%	&\textcolor[rgb]{0.00,0.00,1.00}{26.11} &-8.77\%	&\textcolor[rgb]{0.00,0.00,1.00}{33.32}    &-12.06\%	\\
&WDVSc	&18.73 &-38.83\%	&19.70 &-16.02\%	&30.82 &-19.15\%	\\
&Hyperbolic-ZSL	&\textcolor[rgb]{0.00,0.00,1.00}{25.47} &-3.38\% &21.04	&-38.44\%	&\textcolor[rgb]{0.00,0.00,1.00}{34.49} &-3.66\%	\\
&\cellcolor{LightCyan}\myalgPro\  (ours) &\cellcolor{LightCyan}\textcolor[rgb]{0.00,0.00,1.00}{32.55} &\cellcolor{LightCyan}-3.64\% 	&\cellcolor{LightCyan}\textcolor[rgb]{0.00,0.00,1.00}{31.83} &\cellcolor{LightCyan}-16.28\% 	&\cellcolor{LightCyan}\textcolor[rgb]{0.00,0.00,1.00}{35.05} &\cellcolor{LightCyan}\cellcolor{LightCyan}-16.51\% \\
\bottomrule
\end{tabular}
\begin{tablenotes}
  \item[*]\footnotesize The results which are better than those of RandomGuess are typeset in \textcolor[rgb]{0.00,0.00,1.00}{blue}.
  \item[] \footnotesize The ``Decl.'' column shows the relative decline, compared to the results in Table~\ref{tab_zsl_acc}.
\end{tablenotes}
\label{tab_zsl_acc_name_csd_compare}
\end{table}

\subsection{Component Analysis in \myalgPro\ (RQ2)}
\subsubsection*{\textbf{TEXT-CSDs v.s. LABEL-CSDs}}
To compare these two kinds of CSDs, we conduct a new ZNC experiment by replacing the TEXT-CSDs used in Section~\ref{subsect_exp_znc} with LABEL-CSDs.
As shown in Table~\ref{tab_zsl_acc_name_csd_compare}, the performance of all methods (including ours) declines significantly, compared to those results in Table~\ref{tab_zsl_acc} where TEXT-CSDs are used.
Moreover, more than half of baselines (around 61.11\%) can only (or cannot even) be comparable to RandomGuess.
This definitely shows the superiority of TEXT-CSDs over LABEL-CSDs, which is also consistent with our quantitative CSDs evaluation experiments in Section~\ref{subsect_csd_exps}.

It is worth noting that, unlike our experiments, in the previously published reports in computer vision and natural language processing, LABEL-CSDs generally could get considerable performance.
The reason may be as follows.
In computer vision, concepts can easily be described by very few words (like class names).
In NLP, as instance features are usually given in plain-text form, researchers usually pre-process them by some word2vec tools, which may facilitate the problem.
Another possible reason is that: the recently proposed BERT-Tiny, which really releases the power of TEXT-CSDs, is much better than traditional ones for long text understanding.
We leave this for future study.

\begin{table}[!t]
\setlength{\tabcolsep}{3pt} 
\small
\caption{Zero-shot node classification accuracy (\%) using the graph adjacency information as node attribute information.}
\centering
\begin{tabular}{ll|ccc|ccc}
\toprule
& &\multicolumn{3}{c|}{TEXT-CSDs} &\multicolumn{3}{c}{LABEL-CSDs}\\
& & Cora & Citeseer & C-M10M & Cora & Citeseer & C-M10M \\
\hline
\multirow{6}{*}{\begin{turn}{90}\textbf{Class Split I}\end{turn}}
&DAP &\textcolor[rgb]{0.00,0.00,1.00}{30.76}  &\textcolor[rgb]{0.00,0.00,1.00}{33.98}  &\textcolor[rgb]{0.00,0.00,1.00}{36.76} &\textcolor[rgb]{0.00,0.00,1.00}{28.57} &19.38 &30.91 \\
&ESZSL &24.98 &\textcolor[rgb]{0.00,0.00,1.00}{33.20} &\textcolor[rgb]{0.00,0.00,1.00}{36.34} &\textcolor[rgb]{0.00,0.00,1.00}{30.22} &\textcolor[rgb]{0.00,0.00,1.00}{30.05} &\textcolor[rgb]{0.00,0.00,1.00}{34.61} \\
&ZS-GCN &\textcolor[rgb]{0.00,0.00,1.00}{28.43} &\textcolor[rgb]{0.00,0.00,1.00}{33.35} &\textcolor[rgb]{0.00,0.00,1.00}{36.87} &23.26 &\textcolor[rgb]{0.00,0.00,1.00}{30.26} &\textcolor[rgb]{0.00,0.00,1.00}{33.90} \\
&WDVSc &18.98 &\textcolor[rgb]{0.00,0.00,1.00}{28.77} &\textcolor[rgb]{0.00,0.00,1.00}{33.84} &\textcolor[rgb]{0.00,0.00,1.00}{29.73} &23.03 &30.35 \\
&Hyperbolic-ZSL &19.96 &12.16 &\textcolor[rgb]{0.00,0.00,1.00}{35.80} &\textcolor[rgb]{0.00,0.00,1.00}{28.53} &12.45 &30.82 \\
&\cellcolor{LightCyan}\myalgPro\ (ours) &\cellcolor{LightCyan}\textcolor[rgb]{0.00,0.00,1.00}{32.96} &\cellcolor{LightCyan}\textcolor[rgb]{0.00,0.00,1.00}{38.03} &\cellcolor{LightCyan}\textcolor[rgb]{0.00,0.00,1.00}{40.01} &\cellcolor{LightCyan}\textcolor[rgb]{0.00,0.00,1.00}{31.28} &\cellcolor{LightCyan}\textcolor[rgb]{0.00,0.00,1.00}{31.85} &\cellcolor{LightCyan}\textcolor[rgb]{0.00,0.00,1.00}{35.75} \\
\bottomrule
\end{tabular}
\begin{tablenotes}
  \item[*]\footnotesize The results which are better than those of RandomGuess are typeset in \textcolor[rgb]{0.00,0.00,1.00}{blue}.
\end{tablenotes}
\label{tab_zsl_only_adj}
\end{table}

\subsubsection*{\textbf{Graph Structure v.s. Node Attributes}}
To compare their effects, we take the graph adjacency matrix $A$ as the input node attribute matrix $X$.
Table~\ref{tab_zsl_only_adj} shows the performance with both LABEL-CSDs and TEXT-CSDs.
First, we can see that the results are worse than those in Table~\ref{tab_zsl_acc} where node attributes are used as the input $X$.
This indicates node attributes contain richer and more useful information than graph structure information.
On the other hand, we can see that even with the same input graph adjacency information, the results with TEXT-CSDs are much better than those with LABEL-CSDs.
Specifically, most methods (around 77.78\%) successfully beat RandomGuess in the first case, but fail in the second case.
Especially, on Citeseer with TEXT-CSDs, our method and some compared baselines even could get comparable performance to those in Table~\ref{tab_zsl_acc}.
These observations indicate that the quality of CSDs is the key to ZNC.

\subsubsection*{\textbf{Ablation Study}}
\begin{figure}[!t]
\centering
\includegraphics[width=0.40\textwidth]{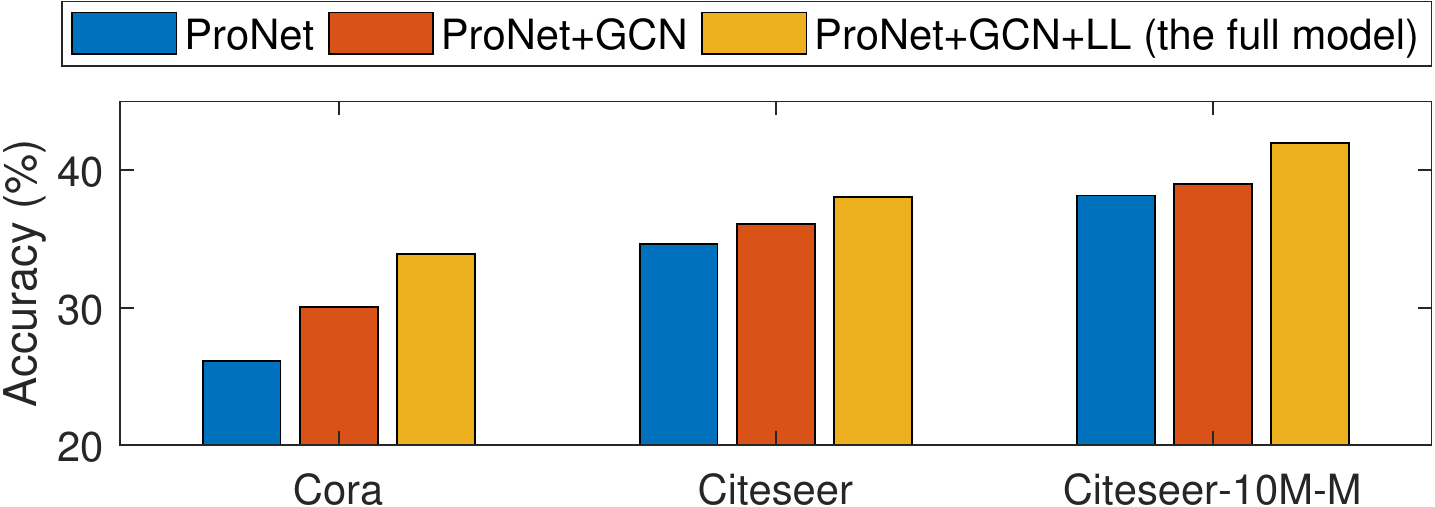}
\caption{Model ablation under Class Split I.}
\label{fig_ablation}
\end{figure}

We test the following three variants of our method:
\begin{itemize}
  \item \emph{ProNet} refers to the variant that replaces the decomposed GCNs part (together with the involved local loss part) with a fully-connected layer.
  This variant can be seen as a classical prototypical network model~\cite{snell2017prototypical}.
  \item \emph{ProNet+GCN} refers to the variant that removes the local loss part in our method.
  This variant can be seen as a special prototypical network which utilizes GCNs as the encoder for node representation learning.
  \item \emph{ProNet+GCN+LL} refers to the exact full model.
\end{itemize}
Figure~\ref{fig_ablation} shows the results of this ablation study.
We can clearly see that both two parts (the decomposed GCNs part and local loss part) contribute to the final performance, which evidently demonstrates their effectiveness.

\begin{figure}[!t]

\centering
\subfigure[Cora]{
    \includegraphics[width=0.147\textwidth]{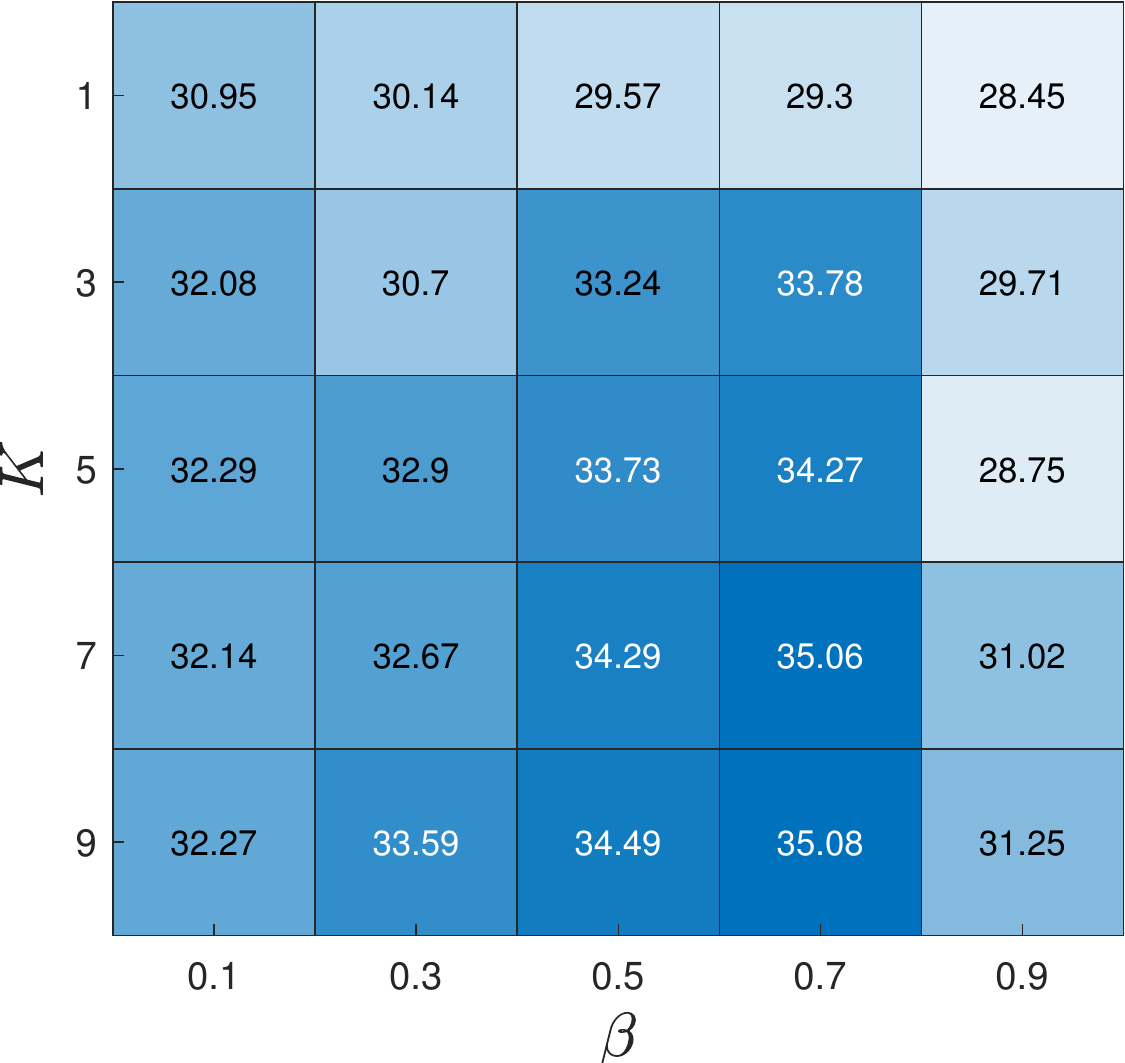}
}
\subfigure[Citeseer]{
    \includegraphics[width=0.147\textwidth]{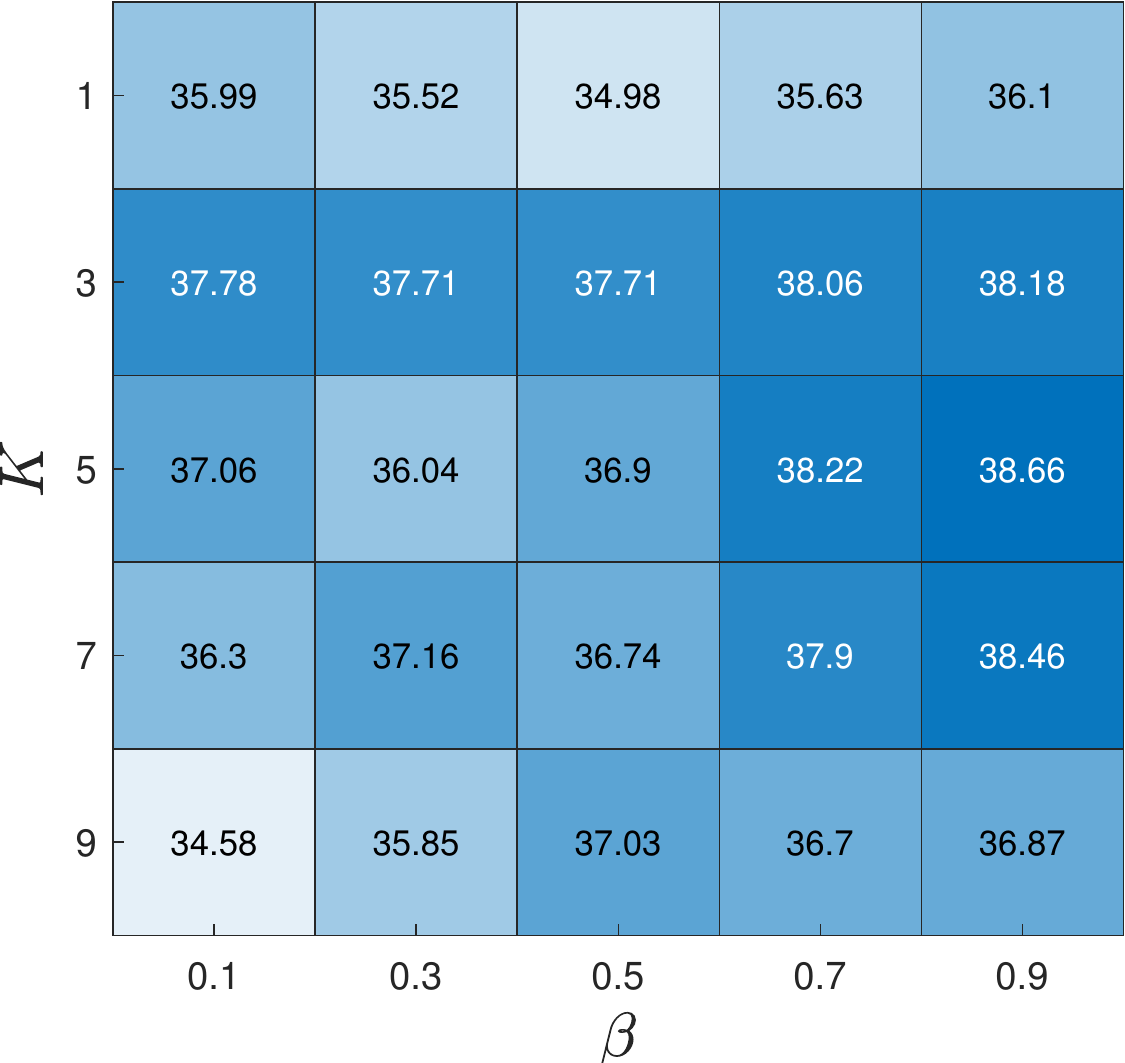}
}
\subfigure[C-M10M]{
    \includegraphics[width=0.147\textwidth]{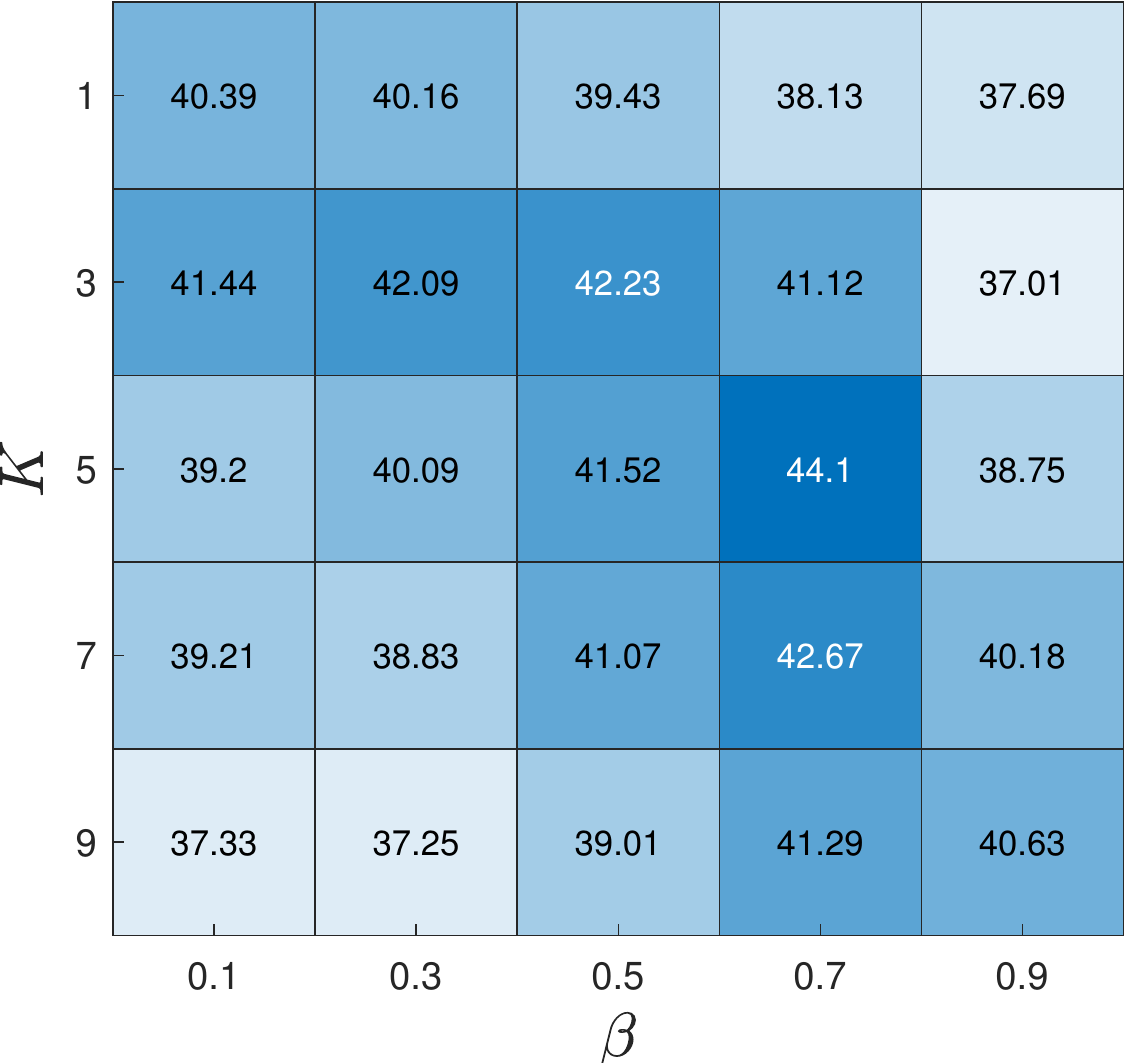}
}
\caption{Effects of $K$ and $\beta$ in our method under Class Split I.
Grid numbers denote the classification accuracy (\%).
Color indicates the performance (the deeper the better).}
\label{fig_k_beta}
\end{figure}

\subsubsection*{\textbf{Parameter Sensitivity}}
Figure~\ref{fig_k_beta} illustrates the performance of our method when the hyper-parameters $K$ and $\beta$ change.
On the whole, when the neighbor hop number $K$ ranges in $[2,3]$ and the $\beta$ ranges in $[0.5, 0.7]$, our method gets the best performance.
These results show the usefulness of the graph structure information and the lazy random walk strategy.

\emph{In sum, we can get some interesting findings: (1) the quality of CSDs is the key to ZNC, (2) with high-quality CSDs, graph structure information can be very useful (even be comparable to node attributes) for ZNC, and (3) the involved decomposed GCNs and local loss play very important roles in our method.}

\subsection{Validation of Decomposed GCNs (RQ3)}
We compare our method with the vanilla GCN method on the standard node classification task.
In our method, we continue with its default network setting, but replace the final prototypical loss with the standard cross-entropy loss.
In addition, we use our trick version (Eq.~\ref{eq_vanila_gcn_trick_decompose_final}) and set $K=2$, so as to be consistent with the default setting of the vanilla GCN.
Besides Cora and Citeseer, we further introduce another citation network Pubmed which has $19,717$ nodes (with $500$-dimensional attributes) and $44,338$ edges.
On all these datasets, we adopt the standard train/val/test splits~\cite{yang2016icml}.
More experimental details can be found in Appendix~\ref{ap_sect_exp_de_gcns}.

Table~\ref{tab_citation_standard_clf} reports the results on the standard node classification task.
We can see that our method obtains very similar results compared to the vanilla GCN method.
This is consistent with our decomposition analysis in Section~\ref{subsubsect_gcn_de_trick}.
Here, we do not test the case with local loss, as this loss does not affect the performance of our method in this task.

\begin{table}[!t]
        \small
        \caption{The standard node classification accuracy (\%).}
        \begin{tabular}{l|c|c|c}
        \toprule
         & Cora & Citeseer & Pubmed \\
        \midrule
        \multicolumn{4}{l}{\textbf{Numbers from literature:}} \\
        GCN  & $81.5$ & $70.3$ & $79.0$  \\
         \midrule
        \multicolumn{4}{l}{\textbf{Our experiments:}} \\
        GCN & $81.23 \pm{0.52}$ & $70.32\pm 0.54$ & $79.01 \pm{0.44}$ \\
        \rowcolor{LightCyan}
        Ours & $81.38 \pm{0.66}$ & $70.53 \pm{0.65}$ &  $78.89 \pm{0.41}$ \\
         \bottomrule
        \end{tabular}
\label{tab_citation_standard_clf}
\end{table}

\emph{In sum, the above experiments demonstrate the effectiveness of our decomposed GCNs part.
This provides new opportunities for our method to be applied in a wider range of applications.} 

%% file: 7.conclusion.tex
\section{Conclusion}\label{sect_conclusion}
In this paper, we  provide the first study of zero-shot node classification.
Our contributions lie in two main points.
First of all, by introducing a novel quantitative CSDs evaluation strategy, we show how to acquire high-quality CSDs in a completely automatic way.
On the other hand, we propose a novel method named DGPN for the studied problem, following the principles of locality and compositionality.
Experiments on several real-world datasets demonstrate the effectiveness of our two main contributions.
In the future, we plan to consider more complex graphs, such as signed graphs and heterogeneous graphs.

%% file: 8.appendix.tex
\begin{table}[!t]
\caption{Detailed class distributions in datasets}
\small
\centering
\begin{tabular}{c|ccc}
\toprule
Dataset & Class ID          & Quantity & Class Label (Name) \\
\midrule
\multirow{7}{*}{Cora}
&0 &818 &Neural Network  \\
&1 &180 &Rule Learning  \\
&2 &217 &Reinforcement Learning \\
&3 &426 &Probabilistic Methods \\
&4 &351 &Theory \\
&5 &418 &Genetic Algorithms \\
&6 &298 &Case Based \\
\midrule
\multirow{6}{*}{Citeseer}
&0 &596 &Agent \\
&1 &668 &Information Retrieval \\
&2 &701 &Database \\
&3 &249 &Artificial Intelligence \\
&4 &508 &Human Computer Interaction \\
&5 &590 &Machine Learning \\
\midrule
\multirow{6}{*}{C-M10M}
&0 &825 &Biology \\
&1 &852 &Computer Science \\
&2 &600 &Financial Economics \\
&3 &730 &Industrial Engineering \\
&4 &674 &Physics \\
&5 &783 &Social Science \\
\bottomrule
\end{tabular}
\label{tab_dataset_classes}
\end{table}

\section{Appendix}

\subsection{Datasets Details}\label{app_data_details}
As summarized in Table~\ref{tab_datasets} and Table~\ref{tab_dataset_classes}, we use the following three real-world datasets:
\begin{enumerate}
    \item \emph{Cora}\footnote{https://linqs-data.soe.ucsc.edu/public/lbc/cora.tgz}~\cite{getoor2005link} is a paper citation network. It consists of 2,708 papers from seven machine learning related categories, with 5,429 citation links among them.
         Each node has a 1,433-dimensional bag-of-words (BOW) feature vector indicating whether each word in the vocabulary is present (indicated by 1) or absent (indicated by 0) in the paper.
    \item \emph{Citeseer}\footnote{https://linqs-data.soe.ucsc.edu/public/lbc/citeseer.tgz}~\cite{getoor2005link} is also a citation network which is a subset of the papers selected from the CiteSeer digital library. It contains 3,312 papers from six categories, with 4,732 citation connections.
        Each node also has a BOW feature vector and the dictionary size is 3,703.
    \item \emph{C-M10M}~\cite{Lang95} is a subset of the scientific publication dataset Citeseer-M10\footnote{https://github.com/shiruipan/TriDNR}~\cite{pan2016tri}.
    As the original Citeseer-M10 contains too much noise, we thereby remove all the nodes which have no labels or attributes, remove the classes whose node number is less than 70, and also remove the associated edges under the above conditions. Finally, we get the dataset \emph{C-M10M} which consists of publications from six distinct research areas, including 4,464 publications and 5,804 citation links. As its node attributes are in plain-text form, we simply use Bert-Tiny to process them to get 128-dimensional features.
\end{enumerate}
\subsubsection*{\textbf{Seen/Unseen Class Split}}
In this paper, we provide two fixed seen/unseen class split settings.
Specifically, based on their class IDs shown in Table~\ref{tab_dataset_classes}, we adopt the first few classes as seen classes and adopt the rest as unseen classes.
In the setting of Class Split I, the [train/val/test] class split for Cora, Citeseer, and C-M10M are: [3/0/4], [2/0/4] and [3/0/3]. In the setting of Class Split II, we further partition the seen classes to train and validation parts, where [train/val/test] class splits in these three datasets become: [2/2/3], [2/2/2] and [2/2/2].

\subsection{CSDs Evaluation Experiment Details}\label{app_csds_exp_details}
In this experiment, we use three real-world datasets: Cora, Citeseer and C-M10M.
The details of these datasets can be found in Appendix~\ref{app_data_details}.
Their node attributes are pre-processed as follows.
As the first two datasets have BOW features, to avoid the curse-of-dimensionality, we apply SVD decomposition to reduce the attribute dimension to 128.
In the third dataset C-M10M, we use the 128-dimensional features generated by Bert-Tiny, as mentioned above.
Finally, all the CSD vectors and node attribute vectors are normalized to unit length, for a fair computation.

\subsection{Explaining Local and Global Features}\label{app_explain_lg}
\begin{figure}[!h]
\centering
\subfigure[Apply CNNs to an image: for this image, the local feature refers to the representations learned from a ``patch'' of an image, and the global feature refers to the pooling (like concatenation) result of those local ones.]{
    \includegraphics[width=0.45\textwidth]{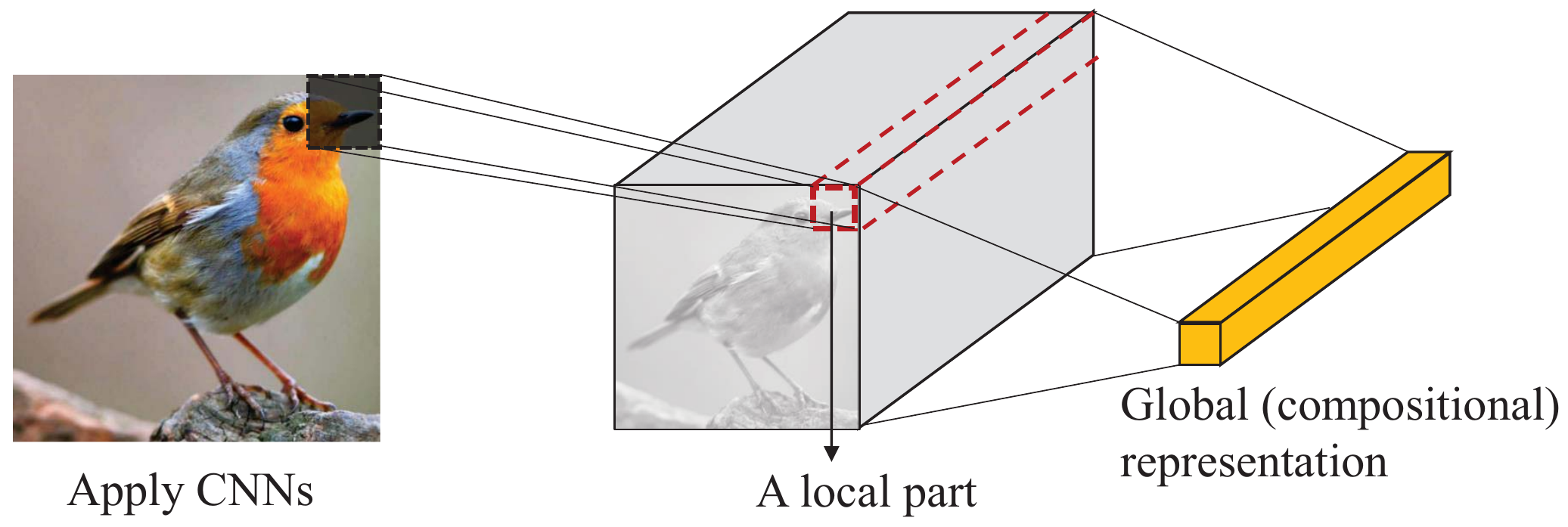}
}
\subfigure[
Apply $K$-layer (here $K{=}2$) GCNs to a graph: for a node in this graph, the $\{k\}_{k=0}^{K}$-th local feature of this node refers to its $k$-hop neighbor information, and the global feature refers to the weighted sum of all its $3$ (i.e., $K+1$) local ones.]{
    \includegraphics[width=0.45\textwidth]{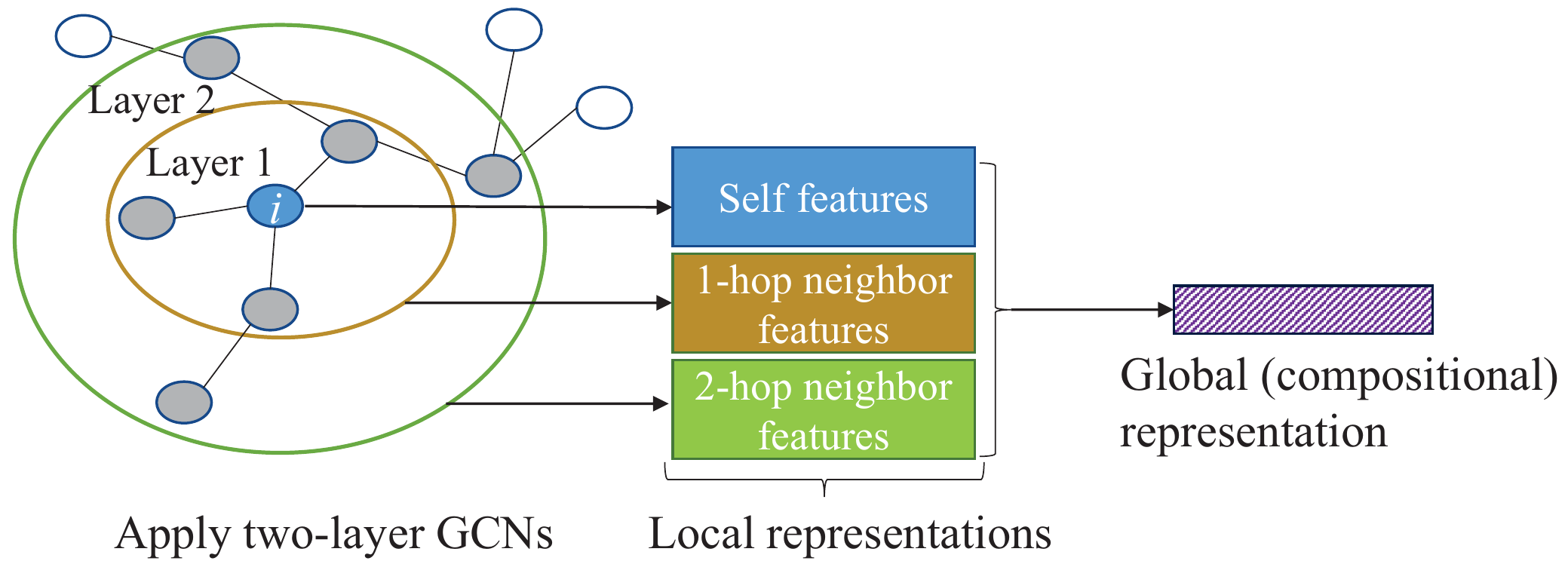}
}
\caption{CNNs for images v.s. GCNs for graphs}
\label{fig_local_global}
\end{figure}

\subsection{Baselines Details}\label{app_baselines}
We compare the result of ours against the following methods:
\begin{enumerate}
    \item \emph{DAP}\footnote{As its codes are unavailable now, we reimplement it in PyTorch.}~\cite{lampert2013attribute} is one of the most well-known ZSL methods.
         In this method, an attribute classifier is first learned from source classes and then is applied to unseen classes for ZSL.
    \item \emph{ESZSL}\footnote{https://github.com/bernard24/Embarrassingly-simple-ZSL}~\cite{romera2015embarrassingly} is a very simple and classical ZSL method. It adopts a bilinear compatibility function to directly model the relationships among features, CSDs and class labels.
    \item \emph{ZS-GCN}\footnote{https://github.com/JudyYe/zero-shot-gcn}~\cite{wang2018zero} uses GCN for knowledge transfer among similar classes, based on class relationships reflected on a knowledge graph.
        However, unlike our method, it applies GCN on a class-level graph which describes the relationships among classes.
    \item \emph{WDVSc}\footnote{https://github.com/raywzy/VSC}~\cite{wan2019transductive} is a transductive ZSL method which jointly considers the samples from both seen and unseen classes.
        It adds different types of visual structural constraints to the prediction results, so as to improve the attribute prediction accuracy.
    \item \emph{Hyperbolic-ZSL}\footnote{https://github.com/ShaoTengLiu/Hyperbolic\_ZSL}~\cite{liu2020hyperbolic} is a recently proposed ZSL method. It learns hierarchical-aware embeddings in hyperbolic space for ZSL, so as to preserve the hierarchical structure of semantic classes in low dimensions.
    \item \emph{RandomGuess} simply randomly choose an unseen class label for each testing node.
\end{enumerate}
In addition, as traditional methods are mainly designed for computer vision, their original implementations heavily rely on some pre-trained CNNs.
Therefore, we additionally test two representative variants: \emph{DAP(CNN)} and \emph{ZS-GCN(CNN)}, in both of which a pre-trained AlexNet is used as the backbone network.
Specifically, we use the AlexNet\footnote{https://pytorch.org/hub/pytorch\_vision\_alexnet/} released by the official PyTorch library.
Also, to be compatible with CNNs, we adopt zero-padding to handle the input of convolution layer.

\subsubsection*{\textbf{Parameter Settings}}
In the Class Split II, the hyper-parameters in baselines and ours are all determined based on their performance on
validation data.
Table~\ref{tab_hyper-parameters} shows the search space of the hyper-parameters.
In addition, in those baselines, their default hyper-parameters are also considered.

\begin{table}[!t]
\small
    \centering
    \caption{Search space of hyper-parameters.}
    \begin{tabular}{c|c}
    Parameters     & Value \\
    \hline
    Learning rate     & \{0.001, 0.01, 0.1\}\\
    Training epoch & \{200, 500, 1000, 1200\} \\
    Weight decay & \{0, 1e-6, 1e-5, 1e-4\} \\
    Dropout rate& \{0.3, 0.5, 0.7\}\\
    K-hop & \{1, 2, 3, 4, 5\}\\
    $\alpha$ & \{0.1, 0.5, 1\}\\
    $\beta$ & \{0.1, 0.3, 0.5, 0.7, 0.9\}\\
    \hline
    \end{tabular}
    \label{tab_hyper-parameters}
\end{table}

\subsection{More Experimental Details of Our Method}\label{app_sect_our_more}
\subsubsection{Running Environment}\label{ap_sect_exp_ent}
The experiments in this paper are all conducted on a single Linux server with 56 Intel(R) Xeon(R) Gold 5120 CPU \@2.20GHz, 256G RAM, and 8 NVIDIA GeForce RTX 2080 Ti.
The codes of our method are all implemented in PyTorch 1.7.0 with CUDA version 10.2, scikit-learn version 0.24, and Python 3.6.

\subsubsection{Detailed Experiments for Decomposed GCNs}\label{ap_sect_exp_de_gcns}
The statistics of the datasets used in this experiment are summarized in Table~\ref{tab_standard_datasets}.
In the vanilla GCN method, we adopt its default setting, i.e., 2 (layer number), 16 (number of hidden units), 0.5 (dropout rate), 5e-4 (L2 regularization), and ReLU (activation function).
We train all methods for a maximum of 200 epochs, using Adam with a learning rate of 0.01.
We also adopt early stopping with a window size of 10, i.e., stopping training if the validation loss does not decrease for 10 consecutive epochs.
In addition, in our method, we also adopt the above training settings.

\begin{table}[tb!]
\small
\centering
\caption{Dataset statistics of the citation networks.}
\label{tab_standard_datasets}
\begin{tabular}{lcccc}
\toprule
Dataset & Classes & Nodes & Edges & Train/Val/Test Nodes \\
\midrule
Cora &7 & $2,708$ & $5,429$ & $140/500/1,000$\\
Citeseer &6 & $3,327$ & $4,732$ & $120/500/1,000$\\
Pubmed &3 & $19,717$ & $44,338$ & $60/500/1,000$\\
\bottomrule
\end{tabular}
\end{table}